%% file: acl_latex.tex
\documentclass[11pt]{article}

\usepackage[preprint]{acl}

\usepackage{times}
\usepackage{latexsym}

\usepackage[T1]{fontenc}

\usepackage[utf8]{inputenc}

\usepackage{microtype}

\usepackage{inconsolata}

\usepackage{graphicx}

\usepackage{makecell}
\usepackage[table]{xcolor} %
\usepackage{booktabs}
\usepackage{amsmath}
\usepackage{amssymb}
\usepackage{enumitem}
\usepackage{algorithm}
\usepackage{algpseudocode}

\title{Not All Attention Heads Contribute to Critical Visual Token Selection: Head-Aware Pruning Matters More}

\author{Chaofang Ma{\mdseries\textsuperscript{1},} Lin Jiang{\mdseries\textsuperscript{2},} Carol Jingyi Li{\mdseries\textsuperscript{1},} Xingyu Liu{\mdseries\textsuperscript{1},} \\
\bfseries Zeyu Li{\mdseries\textsuperscript{1},} Jiang Xu{\mdseries\textsuperscript{3},} {\mdseries and} Wei Zhang{\mdseries\textsuperscript{1*}} \\
  \textsuperscript{1}The Hong Kong University of Science and Technology,
  \textsuperscript{2}Northeastern University, \\
  \textsuperscript{3}The Hong Kong University of Science and Technology (GZ) \\
  \textsuperscript{*}\texttt{Corresponding author: wei.zhang@ust.hk}}

\newcommand{\name}{ProViP}
\definecolor{lightgrays}{gray}{0.95}
\definecolor{mygreen}{RGB}{0, 128, 0}

\begin{document}
\maketitle
\begin{abstract}

Vision–Language Models (VLMs) have exhibited impressive performance across diverse visual scenarios. However, this success comes at the cost of explosive growth in visual tokens, which imposes substantial memory and computational overhead during inference, ultimately increasing latency. To improve VLM inference efficiency, a typical class of visual token pruning methods estimates token importance by aggregating attention scores across all heads in the pruning layer of the Large Language Model (LLM) backbone and prunes tokens based on aggregated scores. However, in this paper, we reveal a compelling phenomenon: the capability to pinpoint critical visual tokens is concentrated within a small fraction of heads.
Aggregation exclusively on these heads can improve task performance.
Inspired by this observation, we propose \name{}, a training-free \textbf{Pro}gressive \textbf{Vi}sual token \textbf{P}runing framework. \name{} first removes redundant visual tokens based on the embedding similarity of input tokens before reasoning of the LLM backbone, and then further prunes tokens during reasoning via head-aware pruning. Experiments demonstrate that \name{} delivers outstanding task performance and inference efficiency. For instance, when applied to LLaVA-1.5-7B, \name{} retains 95.9\% of the original performance and achieves 1.62x inference speedup under an 88.9\% pruning ratio.

\end{abstract}

\input{sections/1_introduction}

\input{sections/2_background}

\input{sections/3_observation}

\input{sections/4_method}

\input{sections/5_evaluation}

\input{sections/6_conclusion}

\section*{Limitations}

While \name{} efficiently accelerates VLM inference, it introduces two primary limitations. First, pruning both before and within the base LLM (including shallow- and middle-layer pruning) requires access to intermediate results during inference. Therefore, our framework is restricted to open-source architectures and cannot be applied directly to closed-source, black-box VLMs (e.g., Gemini (\citealp{team2023gemini, comanici2025gemini}), GPT-5 (\citealp{singh2025openai})), in which such internal states are inaccessible. Second, because our approach requires explicit computation of attention scores at pruning layers in the LLM backbone to guide visual token reduction, it cannot fully bypass attention computation. This introduces overhead that partially offsets the acceleration benefits of high-performance attention implementations, such as FlashAttention \citep{dao2022flashattention, dao2024flashattention}.

\input{acl_latex.bbl}
\clearpage

\appendix

\section{Benchmarks}
\label{sec:appendix}

This section provides a brief overview of each benchmark used in the experiments.

\textbf{GQA (\citealp{hudson2019gqa}).} GQA is a benchmark designed for real-world visual reasoning and compositional question answering. Leveraging Visual Genome scene graph structures, the dataset features 22 million diverse reasoning questions, each accompanied by a functional program that precisely represents its semantics. To ensure evaluation integrity, GQA utilizes a smoothing technique to mitigate inherent question biases. Moreover, it introduces a novel suite of metrics specifically designed to evaluate essential model qualities, such as consistency and grounding.

\textbf{MMB (\citealp{liu2024mmbench})} and \textbf{MMB\(_{\mathrm{CN}}\) (\citealp{liu2024mmbench}).} MMBench is a systematically designed, bilingual objective benchmark developed for a robust and holistic evaluation of VLMs. It addresses the scalability and bias issues of subjective benchmarks, as well as the coarse-grained metrics of traditional ones, by establishing a methodical evaluation pipeline. The benchmark features a larger volume and variety of questions across diverse abilities, curated through strict quality control. To ensure accurate results even for models with limited instruction-following capabilities, MMBench introduces a rigorous CircularEval strategy and leverages LLMs to parse free-form text into predefined choices. Additionally, it offers parallel multiple-choice questions in both English and Chinese, enabling direct, apples-to-apples performance comparisons across a bilingual context.

\textbf{MME (\citealp{fu2023mme}).} MME is the first comprehensive evaluation benchmark designed specifically for VLMs. It simultaneously evaluates both perception and cognition abilities across 14 distinct subtasks. To ensure evaluation integrity and eliminate data leakage risks associated with public datasets, all instruction-answer pairs within MME are manually annotated. Furthermore, the benchmark employs a concise instruction design to enable fair comparisons across various VLMs without the interference of prompt engineering, while also facilitating straightforward quantitative statistics.

\textbf{POPE (\citealp{li2023evaluating}).} POPE is a systematically designed evaluation benchmark specifically developed to investigate and measure the object hallucination problem in VLMs. Addressing the limitations of existing evaluation methods, which are often susceptible to biased input instructions and varying model generation styles, POPE introduces a polling-based query strategy. This formulation converts open-ended description evaluation into a series of closed-ended questions, offering a more stable, flexible, and robust approach to gauging whether an LVLM accurately identifies objects present in a target image.

\textbf{SQA (\citealp{lu2022learn}).} SQA is a large-scale multimodal science question answering benchmark designed to evaluate and interpret the multi-hop reasoning abilities of AI systems. Addressing the limitations of prior datasets, which are often text-only, small-scale, or lack explanatory annotations, SQA comprises approximately 21,000 multimodal multiple-choice questions across a rich diversity of science topics. Crucially, each question is deeply annotated not only with the correct answer but also with its corresponding lecture and detailed explanation. This design explicitly supports chain-of-thought prompting and fine-tuning, allowing models to mimic human-like reasoning by generating structured explanations alongside answers.

\textbf{VQA\(_{\mathrm{Text}}\) (\citealp{singh2019towards}).} VQA\(_{\mathrm{Text}}\) is a specialized visual question answering dataset designed to address the critical need for models to read and reason about text within everyday environments, a dominant requirement for visually impaired users. Bridging the gap left by previous benchmarks, which either contain a negligible proportion of text-related questions or suffer from limited data scales, VQA\(_{\mathrm{Text}}\) provides a large-scale evaluation platform. It comprises 45,336 questions annotated across 28,408 real-world images, specifically curated to ensure that successfully answering the questions strictly necessitates recognizing, interpreting, and reasoning about the scene text in relation to the visual context.

\section{VLM Models}

This section introduces the architectures of the VLM models tested in this paper.

\textbf{LLaVA-1.5-7B (\citealp{liu2024improved}).} LLaVA-1.5-7B is an autoregressive VLM built upon three modular components. For visual perception, it utilizes the CLIP-ViT-L-336px vision encoder with an input resolution of \(336 \times 336\) pixels. The extracted visual tokens are aligned into the language space via a cross-modal connector implemented as a two-layer MLP. The base LLM is Vicuna-7B (\citealp{zheng2023judging}), a decoder-only transformer model with 32 heads per layer across 32 layers. During inference, visual and textual tokens are concatenated along the sequence dimension and processed autoregressively by the base LLM.

\textbf{LLaVA-NEXT-7B (\citealp{liu2024llavanext}).} LLaVA-NeXT-7B enhances the architectural baseline of LLaVA-1.5-7B by introducing a dynamic high-resolution strategy, called AnyRes, to preserve fine-grained visual details. For visual perception, it also uses the CLIP-ViT-L-336px encoder but scales the input resolution up to \(4\times\) more pixels using a dynamic grid configuration (supporting up to \(672 \times 672\), \(336 \times 1344\), or \(1344 \times 336\) based on the image aspect ratio). An image is split into a maximum of 4 patches plus a downsampled global overview patch, with each patch yielding 576 tokens. The visual tokens from all patches are mapped via a two-layer MLP connector and concatenated along the sequence dimension. The fused multi-patch visual embeddings are then processed autoregressively by the LLM backbone.

\textbf{Qwen2.5-VL-7B (\citealp{Qwen2.5-VL}).} Qwen2.5-VL-7B is an upgraded VLM consisting of a dynamic visual encoder and a decoder-only LLM backbone. For visual perception, it utilizes a vision transformer (\citealp{dosovitskiy2020image}) that natively supports arbitrary input resolutions and aspect ratios, extracting visual patches without forced resizing. These dynamic visual features are spatially downsampled and aligned into the language space via an MLP-based connector. The LLM backbone is Qwen2.5-7B (\citealp{qwen2.5}), which integrates multimodal rotary position embedding to seamlessly unify textual, 2D spatial, and 3D temporal positional information for joint autoregressive multimodal reasoning.

\section{Visual Token Pruning Baselines}

This section presents the main ideas of the evaluated baselines in this paper.

\textbf{FastV (\citealp{chen2024image}).} FastV is a training-free visual token pruning method. It leverages attention maps in the shallow layers to evaluate the importance of visual tokens, and subsequently discards unimportant redundant tokens in deep layers to accelerate computation.

\textbf{LLaVA-PruMerge (\citealp{shang2025llava}).} LLaVA-PruMerge first determines token importance by exploiting the attention scores between the \texttt{CLS} token and other visual tokens in the visual encoder to dynamically select the most crucial tokens. Subsequently, to prevent information loss, it clusters the unselected tokens based on key similarity and merges them into the retained visual tokens.

\textbf{MustDrop (\citealp{liu2024multi}).} MustDrop is a multi-stage visual token compression framework that manages token redundancy across the entire model lifecycle: it merges highly similar adjacent tokens and locks a protected key token set in the visual encoder, filters text-irrelevant visual tokens via a dual-attention strategy in the prefilling stage.

\textbf{PDrop (\citealp{xing2024pyramiddrop}).} PDrop is a multi-stage visual token reduction strategy. By partitioning layers of the LLM backbone into sequential stages, it dynamically discards a predefined percentage of visual tokens at the end of each stage based on a lightweight similarity calculation.

\textbf{HiRED (\citealp{arif2025hired}).} HiRED aims to optimize the deployment of high-resolution VLMs. It utilizes attention scores of the \texttt{CLS} token from the vision encoder to dynamically evaluate the information density across different image partitions. Based on this evaluation, it adaptively allocates the token budget to each partition and prunes redundant visual tokens, passing only the most informative tokens to the LLM backbone.

\textbf{VisionZip (\citealp{yang2025visionzip}).} VisionZip is a token selection framework that is training-free to eliminate the redundancy in the outputs of vision encoders. It optimizes the dense visual features right after the vision encoding stage, dynamically extracting only a core subset of informative tokens before they enter the LLM backbone.

\begin{figure*}[t]
  \centering
  \includegraphics[scale=1]{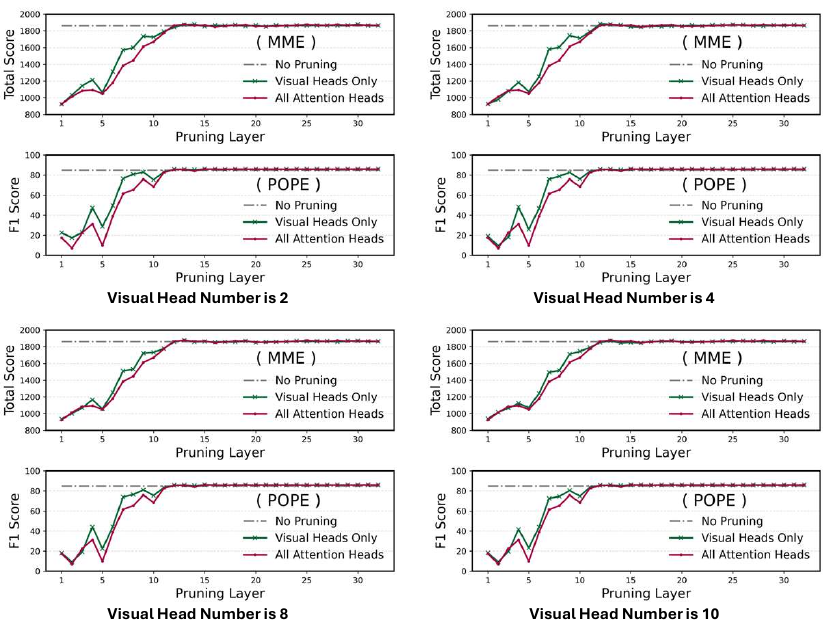}
  \caption{Task performance of layer-wise experiments on LLaVA-1.5-7B across MME and POPE under different settings of visual head numbers. Each layer in the LLM backbone is treated as the pruning layer in turn.}
  \label{fig:obs_more}
\end{figure*}

\textbf{SparseVLM (\citealp{zhang2025sparsevlm}).} SparseVLM is a text-guided token optimization mechanism that dynamically eliminates visual redundancy during inference. It uses attention maps to measure how visual tokens relate to relevant text tokens and then prunes less informative visual features. Furthermore, it incorporates a rank-based strategy to adaptively set the pruning ratio for each layer, alongside a token recycling method that compresses the discarded tokens into a compact representation to retain essential information.

\textbf{DART (\citealp{wen2025stop}).} DART prunes visual tokens based on redundancy rather than traditional importance metrics. It operates by selecting a small subset of pivot tokens and calculating the duplication levels of all other visual tokens relative to these pivots. Subsequently, DART discards highly duplicated tokens and retains those with low duplication.

\textbf{HoloV (\citealp{zoudon}).} HoloV adaptively distributes a predefined pruning budget across different spatial crops of the image. This spatial distribution ensures that the retained tokens capture the entire global visual context rather than isolated salient regions, maintaining robust performance even under aggressive high-pruning ratios.

\textbf{ApET (\citealp{ma2026apet}).} ApET approaches compression from an information-theoretic perspective by using a small set of basis tokens to reconstruct the original visual sequence via linear approximation. The framework then calculates the reconstruction error for each token, using this approximation error as a metric to identify and drop the least informative visual tokens.

\section{Implementation Details}

All experiments in the paper are conducted on the NVIDIA A40 GPU with 40GB of memory. The Python version is 3.10, the PyTorch version is 2.5.1, and the CUDA version is 12.1. The Transformers library (\citealp{wolf2020transformers}) version is 4.37.0 for evaluating LLaVA-1.5-7B and LLaVA-NEXT-7B, and 4.55.4 for evaluating Qwen2.5-VL-7B. Moreover, LMMs-Eval (\citealp{zhang2025lmms}), a unified and standardized multimodal benchmark framework, is leveraged to evaluate task performance on Qwen2.5-VL-7B.

\begin{algorithm}[t]
\caption{Layer-Wise Pruning Evaluation}
\label{alg:layerwise}
\begin{algorithmic}[1]
\Require VLM: LLaVA-1.5-7B; Benchmark set \(\mathcal{B}\): \{MME, POPE\};  
    Configuration set \(\mathcal{C}\): \{\textit{Visual Heads Only}, \textit{All Attention Heads}\}
\For{benchmark \(b \in \mathcal{B}\)}
    \For{\(l = 1\) to \(32\)}
        \For{configuration \(c \in \mathcal{C}\)}
            \State Initialize VLM
            \State Set configuration to \(c\)
            \State Prune at layer \(l\) (576 \(\rightarrow\) 16 tokens)
            \State Evaluate on benchmark \(b\)
            \State Record task performance
        \EndFor
    \EndFor
\EndFor
\end{algorithmic}
\end{algorithm}

\section{Observation with Various Visual Heads}\label{obs_more}

In Section \ref{3}, the impact of visual and non-visual heads is explored when the number of visual heads is set to 6. To provide a more comprehensive analysis, we present evaluation results under different visual head settings. The number of visual heads is set to 2, 4, 8, and 10, respectively. As displayed in Figure \ref{fig:obs_more}, the performance gap between \textit{Visual Heads Only} and \textit{All Attention Heads} across different numbers of visual heads follows the same trend observed in Figure \ref{fig:observe}, gradually narrowing until the two become nearly identical. Notably, as the number of visual heads increases, the performance gap in the same shallow pruning layer narrows. This suggests that when more heads are identified as visual heads, noise is introduced into visual token importance estimation. Hence, this section further verifies the observation in Section \ref{3}: the capability to pinpoint critical visual tokens is concentrated within a small fraction of heads.

\section{Pseudo-Code of Section \ref{3}}\label{pse}

Algorithm \ref{alg:layerwise} presents the deployment details of the layer-wise experiments described in Section \ref{3}.

\section{Computational Cost}

\subsection{Cost without Pruning}
Since the base LLM accounts for the majority of computation in VLM inference, we primarily analyze the computational cost (FLOPs) during LLM reasoning. Moreover, as the number of visual tokens significantly exceeds that of textual tokens, we focus on the computational cost associated with visual tokens. Specifically, within each layer of the LLM backbone, which consists of \(N_h\) heads, the main computational workloads are Multi-Head Attention (MHA) and Feed-Forward Network (FFN).

We first measure the computational cost without pruning. We assume that the original input visual token number of the LLM backbone is \(N_v\), the embedding dimension is \(d\), and the intermediate state size of FFN is \(4d\). For each layer, the computational cost of MHA and FFN is \(8N_vd^2 + 4N_v^2d\) and \(16N_vd^2\), respectively. Overall, the computational cost of all layers is calculated by
\begin{equation}
       T_{vanilla}  \approx 24LN_vd^2 + 4LN_v^2d,
\end{equation}
where \(L\) is the layer number of the base LLM. 

\subsection{Cost with \name{}}
For the VLM pruned by \name{}, the computational cost consists of two components: (1) the cost of MHA and FFN operations, and (2) the cost of retaining task-relevant visual tokens. To better formalize, we assume that before LLM reasoning, the number of visual tokens is first reduced from the original \(N_v\) to \(N_1\) through pruning before the base LLM. During LLM reasoning, the number of visual tokens is further reduced to \(N_2\) at the \(P_i\)-th layer and \(N_3\) at the \(P_j\)-th layer, where \(P_j > P_i\). 

The first step is to analyze the total computational cost of MHA and FFN. The cost from the first layer to \(P_i\)-th layer is computed by
\begin{equation}
  T_1 \approx 24P_iN_1d^2 + 4P_iN_1^2d.
\end{equation}
Similarly, the cost from the (\(P_i+1\))-th layer to the \(P_j\)-th layer is computed by
\begin{equation}
  T_2 \approx 24(P_j-P_i)N_2d^2 + 4(P_j-P_i)N_2^2d.
\end{equation}
The cost from the (\(P_j\)+1)-th layer to \(L\)-th layer is computed by
\begin{equation}
  T_3 \approx 24(L-P_j)N_3d^2  + 4(L-P_j)N_3^2d.
\end{equation}
Overall, the total computational cost of MHA and FFN when applying \name{} is
\begin{equation}
\begin{aligned}
      T_{MF} \approx &24(P_iN_1+(P_j-P_i)N_2 \\
             & +(L-P_j)N_3)d^2 + 4(P_iN_1^2 \\
             & +(P_j-P_i)N_2^2+(L-P_j)N_3^2)d.
\end{aligned}
\end{equation}

Subsequently, the cost of retaining task-relevant visual tokens for each stage is analyzed. Since the critical textual token selection stage only refers to textual tokens, the overhead is negligible. For the pruning before the base LLM, the cost is 
\begin{equation}
    T_4 \approx 2d(|\mathcal{T}|N_v + (N_v-N_1)N_1).
\end{equation}
For the shallow-layer pruning, the cost is formulated as
\begin{equation}
    T_5 \approx 3N_h|\mathcal{T}|N_1 + 6N_1|\mathcal{T}|,
\end{equation}
and the cost of the middle-layer pruning is
\begin{equation}
    T_6 \approx 32N_2|\mathcal{T}|.
\end{equation}
Given that \(|\mathcal{T}|\) is small, the overall cost of retaining task-relevant visual tokens is about
\begin{equation}
    T_{RT} \approx 2d(N_v-N_1)N_1.
\end{equation}

In summary, the total computational cost with \name{} is computed by
\begin{equation}
    T_{\name{}} = T_{MF}+T_{RT}.
\end{equation}
The computational cost reduction ratio is then calculated by
\begin{equation}
    1-\frac{T_{\name{}}}{T_{vanilla}}.
\end{equation}

\begin{table*}[t] 
\centering
\setlength{\aboverulesep}{0pt} %
\setlength{\belowrulesep}{0pt} %
\small

\begin{tabular}{l|ccccccc|c}
\toprule
\textbf{Method} & \textbf{GQA} & \textbf{MMB} & \textbf{MMB\(_{\mathrm{CN}}\)} & \textbf{MME} & \textbf{POPE} & \textbf{SQA}  & \textbf{VQA\(_{\mathrm{Text}}\)} &  \textbf{Average}  \\
\midrule
\rowcolor{lightgrays} LLaVA-1.5-7B & \multicolumn{8}{c}{\textit{Original 576 Tokens }} \\ 
Vanilla  & 61.9 & 64.7 & 58.1 & 1,862 & 85.9 & 69.5 & 58.2 & 100\% \\
\midrule
\rowcolor{lightgrays} LLaVA-1.5-7B &  \multicolumn{8}{c}{\textit{Retain 192 Tokens (\textbf{\(\downarrow\) 66.7\%})}} \\

\rowcolor[HTML]{F3EFFF} \name{} \tiny(Ours)  & 60.9&	64.8&	59.1&	1,824	&85.5&	68.7&	57.4&	\textbf{99.3\%}
 \\

Only Pruning before the Base LLM & 59.4 & 62.2 & 54.7 & 1,803 & 86.8 & 69.0  & 55.9  & 97.0\%  \\

\midrule

\rowcolor{lightgrays} LLaVA-1.5-7B & \multicolumn{8}{c}{\textit{Retain 128 Tokens (\textbf{\(\downarrow\) 77.8\%})}} \\

\rowcolor[HTML]{F3EFFF} \name{} \tiny(Ours)& 60.2&	63.9&	57.8&	1,785&	86.3&	69.2	&56.7&	\textbf{98.4\%}
 \\

Only Pruning before the Base LLM & 58.7 & 62.7 & 53.4 & 1,742 & 86.2 & 69.7  & 54.4 & 95.9\%   \\

\midrule
\rowcolor{lightgrays} LLaVA-1.5-7B & \multicolumn{8}{c}{\textit{Retain 64 Tokens (\textbf{\(\downarrow\) 88.9\%})}} \\

\rowcolor[HTML]{F3EFFF} \name{} \tiny(Ours)& 58.7&	62.3&	54.5&	1,723&	85.8&	69.4&	54.9&	\textbf{95.9\%}
 \\

Only Pruning before the Base LLM & 56.8& 61.3 & 49.1  & 1,694 & 83.5 & 70.2 & 53.0 & 93.0\%  \\

\midrule
\rowcolor{lightgrays} LLaVA-1.5-7B & \multicolumn{8}{c}{\textit{Retain 48 Tokens (\textbf{\(\downarrow\) 91.7\%})}} \\

\rowcolor[HTML]{F3EFFF} \name{} \tiny(Ours) & 57.5&	62.4&	53.4&	1,682&	85.4&	69.6&	53.6&	\textbf{94.7\%}
 \\

Only Pruning before the Base LLM & 56.0 & 59.0 & 47.2  & 1,617 & 82.2 & 70.1 & 51.9 & 90.8\%  \\

\midrule
\rowcolor{lightgrays} LLaVA-1.5-7B & \multicolumn{8}{c}{\textit{Retain 32 Tokens (\textbf{\(\downarrow\) 94.4\%})}} \\

\rowcolor[HTML]{F3EFFF} \name{} \tiny(Ours) & 56.3&	60.3&	51.2&	1,656&	83.9&	69.7&	52.5&	\textbf{92.8\%}
 \\

Only Pruning before the Base LLM & 54.6 & 57.8& 45.1  & 1,531 & 79.7 & 69.6 & 50.0 & 88.0\%  \\

\midrule
\rowcolor{lightgrays} LLaVA-1.5-7B & \multicolumn{8}{c}{\textit{Retain 16 Tokens (\textbf{\(\downarrow\) 97.2\%})}} \\

\rowcolor[HTML]{F3EFFF} \name{} \tiny(Ours) & 53.5&	58.7&	46.0&	1,597&	79.7&	70.2&	50.1&	\textbf{88.9\%}
 \\

Only Pruning before the Base LLM & 51.5 & 53.1 & 38.2 & 1,444 & 73.8 & 68.4 & 46.9 & 81.9\%  \\

\bottomrule
\end{tabular}
\caption{
Task performance on LLaVA-1.5-7B between \name{} and only pruning before the LLM backbone.
}
\label{ablation_2}
\end{table*}

\begin{figure*}[!t]
 \centering
  \includegraphics[width=1\textwidth]{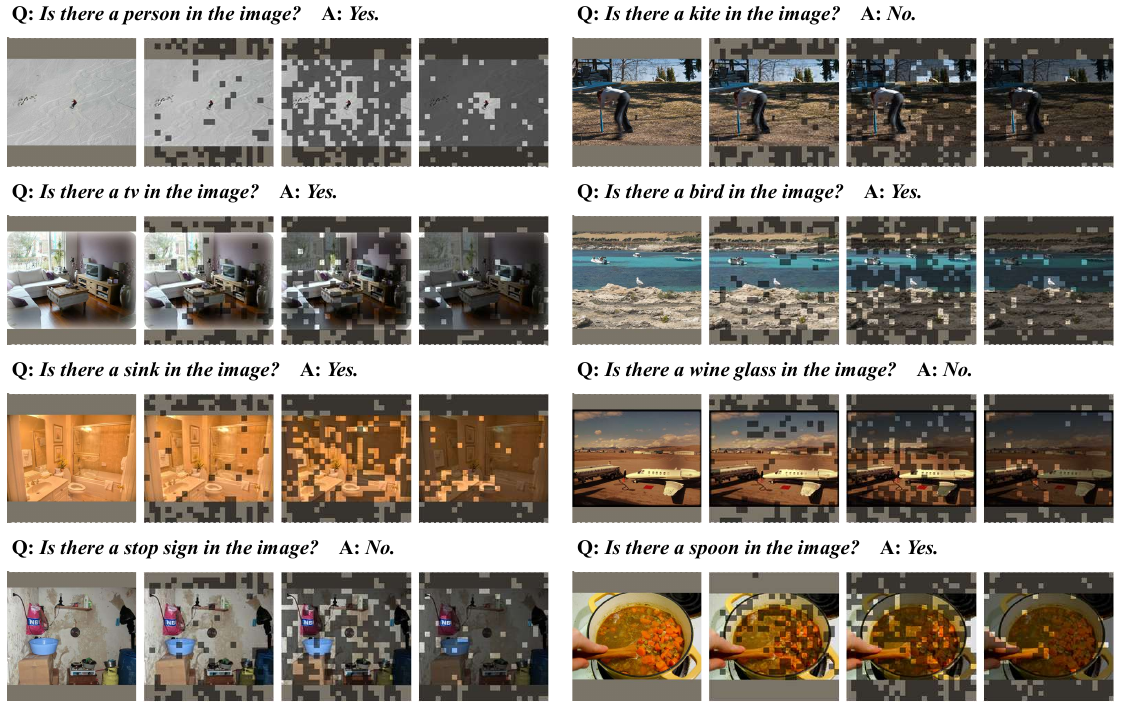}
  \caption{Visualization examples on POPE when applying \name{} to LLaVA-1.5-7B.}
  \label{fig:vis}
\end{figure*}

\section{More Ablation Results} \label{ablation_more}

To further verify the effectiveness of \name{}, we compare the task performance of LLaVA-1.5-7B using only the pruning before the LLM backbone with that of the full \name{} framework. The pruning ratios are set to 66.7\%, 77.8\%, 88.9\%, 91.7\%, 94.4\%, and 97.2\%, respectively. As illustrated in Table \ref{ablation_2}, \name{} consistently outperforms only pruning before the LLM backbone across all pruning ratios. Moreover, the performance gap increases from 2.3\% to 8.0\% as the pruning ratio increases, highlighting the effectiveness of combining all the components in \name{}.

\section{Visualization}

Figure \ref{fig:vis} illustrates examples to intuitively display how \name{} retains task-relevant visual tokens to maintain task performance and enhance efficiency.

\end{document}

%% file: sections/1_introduction.tex
\section{Introduction}

Benefiting from the prosperity of Large Language Models (LLMs) (\citealp{zhao2023survey, naveed2025comprehensive, liu2026systematic}), Vision–Language Models (VLMs) (\citealp{achiam2023gpt, wu2024deepseek, liu2024improved, chen2024internvl, li2025survey}) exhibit impressive capabilities in visual scenes. They have been widely deployed across diverse domains, such as medical analysis (\citealp{li2023llava, wang2025interpretable, nath2025vila}), autonomous driving (\citealp{li2025fine, tian2025nuscenes, tian2025large}), and content moderation (\citealp{guo2024moderating, levi2025ai}). However, despite the remarkable performance, increasing scenario complexity (\citealp{shao2025large, tang2025video, brimont2026survey}) induces a proliferation of visual tokens, imposing substantial memory and computational overhead and ultimately increasing VLM inference latency.

To tackle the challenge, visual token pruning (\citealp{yao2026towards, shaosurvey}) has emerged as one of the promising techniques (\citealp{shinde2025survey, jin2025efficient, kang2026vispec}). This approach is motivated by the inherent redundancy in visual information, as only a small subset of tokens contributes to task performance. By selectively retaining task-relevant visual tokens, pruning reduces memory and computational burdens, thereby improving efficiency while preserving performance.

\begin{figure}[t]
  \centering
  \includegraphics[scale=0.99]{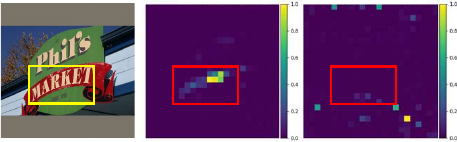}
  \caption{Attention maps of two heads under the textual instruction “What is the word in the red part?” The left head focuses on task-relevant regions, while the right head attends to irrelevant areas. \name{} removes redundant visual tokens via head-aware pruning.}
  \label{fig:motivate}
\end{figure}

Among existing visual token pruning methods, a particularly effective class (\citealp{chen2024image, zhang2025sparsevlm}) utilizes attention maps from the LLM backbone to guide token pruning. In these methods, token importance is quantified by aggregating attention scores, commonly through averaging, across all heads within the pruning layer. Visual tokens with higher importance are identified as critical and retained to maintain performance, while the others are pruned to enhance efficiency.

While aggregating attention scores from all heads provides a convenient strategy, it implicitly assumes that each head contributes equally to token importance estimation. To examine whether this assumption effectively identifies critical visual tokens, we conduct experiments in this paper. Our empirical analysis reveals a compelling phenomenon: the capability to pinpoint critical visual tokens is concentrated in a small fraction of heads. We refer to these visually selective heads as \textit{\textbf{visual heads}}, with the formal description in Section \ref{3}. 
Restricting aggregation on attention scores of visual heads noticeably improves task performance when pruning is applied to shallow layers of the base LLM, whereas the gains diminish in deeper layers.

Inspired by these findings, we propose \name{}, an effective and training-free framework for progressive visual token pruning. \name{} preliminarily reduces visual redundancy before visual tokens enter the LLM backbone and further removes visual tokens within shallow and middle layers via head-aware pruning. By combining them, the framework improves VLM inference efficiency while maintaining competitive task performance.

In summary, our contributions are as follows:

\begin{itemize}

\item We uncover a layer-dependent phenomenon in attention-based visual token pruning: aggregation solely on visual heads yields noticeable performance gains in shallow layers, whereas the gains are less pronounced in deeper layers.

\item Based on this observation, we propose \name{}, a training-free progressive visual token pruning framework that removes redundant tokens before LLM reasoning and performs head-aware pruning during reasoning.

\item Experiments indicate that \name{} achieves outstanding inference efficiency and task performance across a series of benchmarks.

\end{itemize}

%% file: sections/2_background.tex
\section{Related Work}

\textbf{VLM Architecture.} 
In a typical VLM pipeline, raw visual inputs are transformed into visual tokens, which are subsequently concatenated with textual tokens and processed by an LLM backbone (e.g., Vicuna (\citealp{zheng2023judging}), Qwen(\citealp{bai2023qwen}), InternLM2 (\citealp{cai2024internlm2}), Qwen2 (\citealp{yang2407qwen2})) for joint reasoning. Specifically, before the LLM backbone reasoning, a visual encoder (e.g., CLIP (\citealp{radford2021learning}), SigLIP (\citealp{zhai2023sigmoid})) first extracts semantically rich feature representations from the visual inputs. A visual connector, typically an MLP projector (\citealp{lu2024deepseek, chen2024far, wang2024cogvlm, chen2024sharegpt4v}), then maps these representations into the base LLM's input embedding space, ensuring that visual embeddings are aligned with textual counterparts in both dimension and semantics.

\noindent 
\textbf{Visual Token Pruning Methods.} Since not all visual tokens carry task-relevant information, pruning visual tokens improves efficiency with negligible performance degradation. Previous works have proposed methods for removing redundant visual tokens (\citealp{nguyen2025token, shaosurvey}). From the perspective of pruning principles, these methods broadly fall into two classes: attention-based and similarity-based. Attention-based methods
(\citealp{chen2024image, zhang2025sparsevlm}) 
usually evaluate the importance of visual tokens using the base LLM's attention scores, and retain tokens with high importance. However, treating all heads equally in these methods inevitably introduces noise, leading to suboptimal token pruning. Similarity-based methods 
(\citealp{wen2025stop, ma2026apet}) 
use embedding correlations among visual tokens to retain informative tokens. Since these methods ignore textual instructions, they cannot adaptively preserve task-relevant visual tokens.

%% file: sections/3_observation.tex
\section{Observation: Not All Heads Matter} \label{3}

In this section, the empirical findings that motivate \name{} are presented. 
Conventional attention-based methods implicitly assume that all heads contribute equally to identifying critical visual tokens, thereby overlooking the heterogeneous contributions of different heads. We hypothesize that the capability to localize salient visual tokens is concentrated in a small subset of heads (visual heads), whereas the remaining heads introduce redundant or even detrimental noise.
To validate this hypothesis, we introduce a lightweight heuristic to distinguish visual heads from others. By isolating visual heads, this heuristic enables a comprehensive analysis of how token importance indicators constructed from different head groups affect task performance. Empirical findings confirm our hypothesis.

\subsection{Visual Head Selection Heuristic} \label{probe}

The core definition behind our heuristic is that a visual head exhibits sharp semantic focus when processing visual inputs, with its attention highly concentrated on visual tokens relevant to the textual instruction. Conversely, a non-visual head exhibits indiscriminate or misleading visual selectivity, either dispersing its attention across all visual tokens or concentrating on task-irrelevant ones, thereby introducing noise into token importance estimation. Figure \ref{fig:motivate} illustrates a case of the differences in attention focus between a visual head (left) and a non-visual head (right) for better understanding. Driven by this definition, we categorize all the heads within the pruning layer through the following steps (described in detail in Section \ref{4}):

\textbf{Semantic Anchor Extraction.}
Given a textual instruction, textual tokens with the core semantic keyphrases are first extracted to serve as cross-modal semantic guidance anchors. 

\textbf{Confidence Metric.} For each head within the pruning layer, we define the variance of its cross-modal attention scores over the extracted textual tokens as the confidence metric. A larger confidence indicates greater concentration on text-related visual regions, reflecting stronger visual selectivity.

\textbf{Head Classification.} We rank all heads in the pruning layer based on their confidence scores. The heads with top scores are defined as visual heads, while the others are considered non-visual heads.

\subsection{Empirical Findings} \label{insight}

To systematically observe the influence of head selection on task performance, we conduct layer-wise experiments on LLaVA-1.5-7B (\citealp{liu2024improved}) across two prominent benchmarks: MME (\citealp{fu2023mme}) and POPE (\citealp{li2023evaluating}). Specifically, we let each layer in the base LLM take turns serving as the pruning layer. Performance under two configurations is compared: 
(1) \textbf{\textit{All Attention Heads}}, the typical baseline that aggregates attention scores across all heads of the pruning layer, and 
(2) \textbf{\textit{Visual Heads Only}}, which restricts the aggregation solely to the identified visual heads. Empirically, the top 6 out of the 32 heads are identified as visual heads based on the confidence metric (detailed ablation on visual head number is provided in Appendix \ref{obs_more}).
For a stringent setting, we adopt an aggressive pruning strategy, reducing the original 576 visual tokens to merely 16 at the pruning layer. This demands precise token selection, thereby amplifying the performance gap between the two configurations and exposing the impact of visual and non-visual heads.
The above implementation details are summarized in the pseudocode provided in Appendix~\ref{pse}.

\begin{figure}[t]
  \centering
  \includegraphics[scale=0.44]{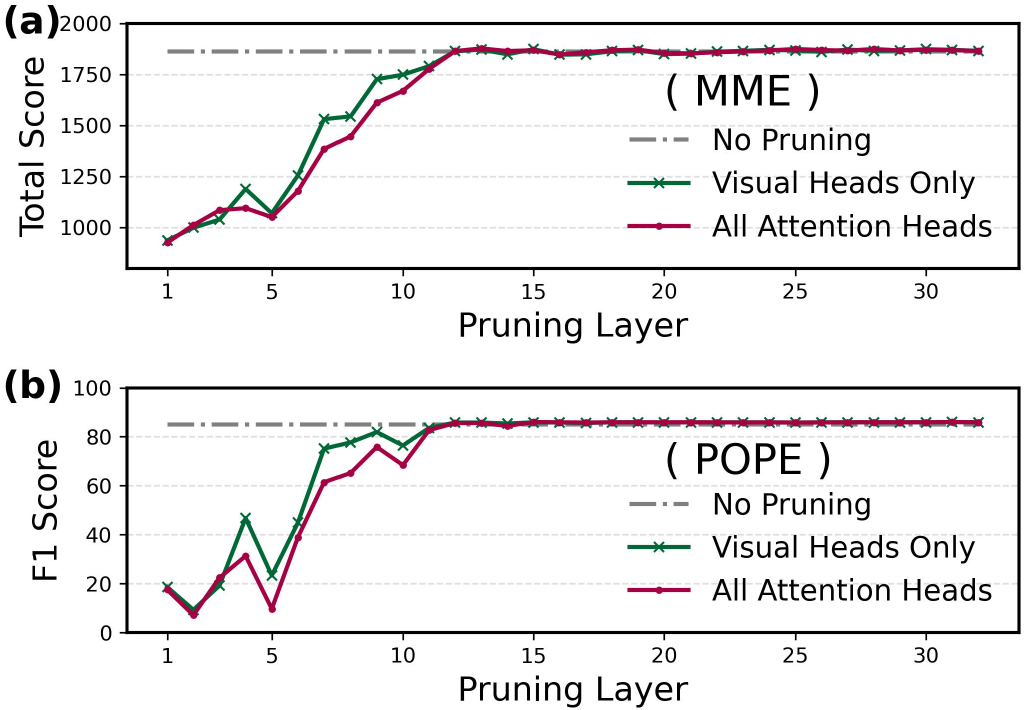}
  \caption{Task performance between \textit{All Attention Heads} and \textit{Visual Heads Only} under different pruning layers: (a) Total score of MME; (b) F1 score of POPE.}
  \label{fig:observe}
\end{figure}

As shown in Figure \ref{fig:observe}, task performance under both configurations exhibits consistent trends across the two benchmarks, initially increasing and then converging. When visual tokens are pruned in the shallow layers, the \textit{Visual Heads Only} configuration outperforms the \textit{All Attention Heads} baseline by an evident margin. For example, the F1 score gap reaches 13.7 when pruning is performed at the 7th layer on POPE. This observation demonstrates that estimating token importance based on the entire head set is polluted by non-visual heads of the LLM backbone. Moreover, it suggests that the proposed heuristic effectively identifies visual heads, resulting in a cleaner importance estimator.

By contrast, when pruning is applied to the middle or deep layers, the performance gap gradually diminishes. 
We attribute this convergence to the fact that visual tokens tend to achieve homogeneity through extensive global cross-modal interactions in these layers (\citealp{nguyen2023mitigating}), rendering the model inherently more robust to potentially inaccurate token selection.
Overall, since redundant visual tokens are typically removed in the shallow layers to maximize efficiency, our findings indicate that pruning guided by \textit{Visual Heads Only} can yield crucial performance gains where it matters most.

\begin{figure*}[t]
 \centering
  \includegraphics[width=0.98\textwidth]{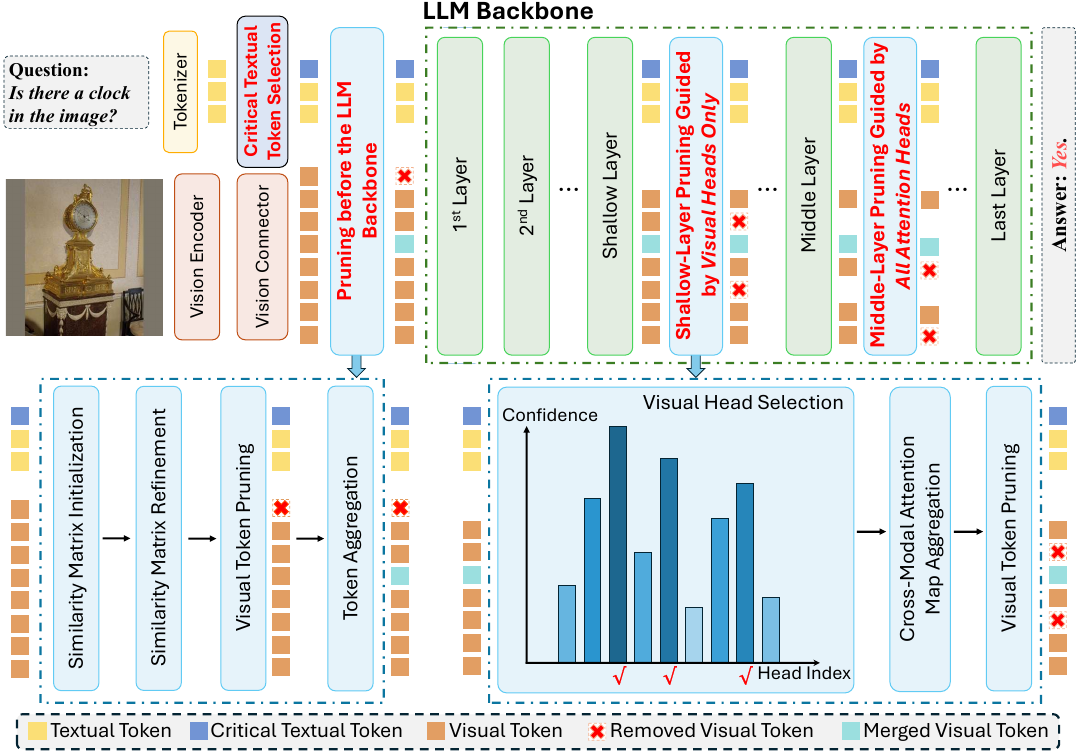}
  \caption{Overview of \name{}. \name{} consists of the progressive stages: critical textual token selection, pruning before the base LLM, and pruning during the base LLM (including shallow-layer and middle-layer pruning). }
  \label{fig:design}
\end{figure*}

%% file: sections/4_method.tex
\section{Method}\label{4}

\subsection{Overview}

As revealed in Section \ref{insight}, the capability to identify task-relevant visual tokens is concentrated in visual heads. Inspired by this, we propose \name{}, a training-free progressive visual token pruning framework to accelerate inference while minimizing performance loss. As Figure \ref{fig:design} shows, \name{} removes redundant tokens via the sequential stages: 

\textbf{Critical Textual Token Selection.} 
To extract the semantic anchors mentioned in Section \ref{probe}, we propose and formalize a strategy for identifying these anchors to guide subsequent pruning.

\textbf{Pruning before the Base LLM.} 
Since attention maps are unavailable before the base LLM, we apply text-guided similarity-based pruning before tokens enter the LLM backbone. By removing redundant visual tokens at this stage, we substantially reduce inference overhead, as pruned tokens are excluded from all subsequent computations.

\textbf{Pruning during the Base LLM.} To maximize efficiency, pruning should be applied to the shallow or middle layers rather than the deep layers, as early sequence reduction shortens the input for most layers, thereby reducing overall overhead. In \name{}, pruning is executed twice sequentially: once in a shallow layer and once in a middle layer. 
Based on the observation that the performance gap between \textit{All Attention Heads} and \textit{Visual Heads Only} is evident in the shallow layers but mild in the middle layers, we adopt \textit{Visual Heads Only} for shallow-layer pruning to retain critical visual tokens effectively. 
The formalization of the confidence metric and head classification in Section \ref{probe} is provided in Section \ref{4.4}.
For middle-layer pruning, we employ \textit{All Attention Heads}, as visual information is sufficiently integrated across heads, enabling the use of more comprehensive attention scores.

\subsection{Critical Textual Token Selection}  \label{4.2}

Since non-content words in textual instructions, such as prepositions, seldom correspond to visual features, it is essential to retain only semantically crucial textual tokens to avoid introducing potential noise to the following pruning stages. To this end, textual self-similarity is used to ensure that the selection of critical textual tokens is decoupled from visual information. This approach retains critical textual tokens, even when they lack direct visual correspondence. Specifically, the process is executed after the tokenizer produces the full sequence of textual tokens. Let the number of textual tokens be \(N_t\). Formally, for textual embeddings \(\boldsymbol{E}_{t}\in \mathbb{R}^{N_t \times d}\), where \(d\) is the embedding dimension, ranking criterion of textual tokens \(\boldsymbol{r}\in \mathbb{R}^{1 \times N_t}\) is

\begin{equation}
    \boldsymbol{r} = \frac{1}{N_t}\sum_{i = 1}^{N_t} \big( \mathrm{Softmax} ( \boldsymbol{E}_{t} \cdot \boldsymbol{E}_{t}^T ) \big)_i.
\end{equation}
Then the index set of critical textual tokens is 
\begin{equation}
\mathcal{T} = \{j\, |\, \boldsymbol{r}_j > \mathrm{Mean}(\boldsymbol{r})\}.
\end{equation}

\subsection{Pruning before the LLM Backbone } \label{4.3}

Before jointly feeding textual and visual tokens into the LLM backbone, a text-guided similarity-based pruning is applied. Specifically, a cross-modal similarity matrix is built and recalibrated to emphasize task-relevant visual tokens. Visual tokens are then prioritized based on their peak similarity to critical textual tokens, ensuring that informative tokens are retained while redundant ones are removed.

We first evaluate the cross-modal relevance by computing the similarity matrix \(\boldsymbol{M}\in \mathbb{R}^{|\mathcal{T}| \times N_v}\), where \(N_v\) is the original number of visual tokens produced by the vision connector. Given \(\boldsymbol{E}_{t}\) and visual embeddings \(\boldsymbol{E}_{v} \in \mathbb{R}^{N_v \times d}\) from the visual connector, \(\boldsymbol{M}_{i,j}\) characterizes the affinity between the \(i\)-th critical textual token and the \(j\)-th visual token via cosine similarity, formulated as

\begin{equation}
\boldsymbol{M}_{i,j} =  \frac{\boldsymbol{E}_{t,t_i} \cdot \boldsymbol{E}_{v,j}^T}{\|\boldsymbol{E}_{t,t_i}\| \|\boldsymbol{E}_{v,j}\|},\, \mathrm{where}\, t_i \in \mathcal{T}.
\end{equation}

Subsequently, the similarity matrix is refined to mitigate the impact of non-discriminative critical textual tokens that exhibit uniformly high affinity scores across all visual tokens. We calculate the mean similarity score for each textual token across all visual tokens, and then subtract it from the original affinity values. This operation suppresses semantic noise and prevents critical textual tokens with high but uniform similarity scores across all visual tokens from dominating the selection. Formally, \(\hat{\boldsymbol{M}}_{i,j}\) is refined by
\begin{equation}
\hat{\boldsymbol{M}}_{i,j} = \boldsymbol{M}_{i,j} - \frac{1}{N_v} \sum_{k=1}^{N_v} \boldsymbol{M}_{i,k}.
\end{equation}
A max-pooling operation is then applied across the textual dimension to derive the importance score for each visual token, formulated as
\begin{equation}
\boldsymbol{p}_j = \max \{ \hat{\boldsymbol{M}}_{1,\,j}, \hat{\boldsymbol{M}}_{2,\,j}, \dots, \hat{\boldsymbol{M}}_{|\mathcal{T}|,\,j} \}.
\end{equation}
This mechanism ensures that a visual token is prioritized as long as it shows a peak similarity score with one or more critical textual tokens. The indices of visual tokens to be retained are finally identified by the top \(N_{1}\) scores according to \(\boldsymbol{p} \in \mathbb{R}^{1 \times N_v} \), while the remaining ones are marked for pruning.

Rather than indiscriminately discarding these unselected visual tokens, our framework treats the retained tokens as cluster centroids to preserve potentially salient information from the pruned counterparts. The embeddings of the pruned tokens are aggregated into their nearest centroids following ApET (\citealp{ma2026apet}). By absorbing these residual embeddings, this approach mitigates information loss from task-relevant visual tokens that may be sub-optimally ranked.

\subsection{Pruning during the LLM Backbone} \label{4.4}

For simplicity, we present only the formulation for shallow-layer pruning under \textit{Visual Heads Only}, since middle-layer pruning under \textit{All Attention Heads} is a simplified variant that expands the candidate pool to all attention heads without the process of visual head identification.

To identify visual heads, a confidence metric is proposed to evaluate each attention head within the pruning layer. According to our definition in Section \ref{probe}, a visual head exhibits sharp semantic focus when processing visual inputs, with its attention highly concentrated on visual tokens most relevant to the textual instruction. For each attention head, we quantify the concentration for each critical textual token by computing the variance of its cross-modal attention scores. These variances are then summed to yield a head-level confidence metric. A larger value indicates a greater likelihood that the head functions as a visual head. Formally, for a cross-modal attention map \(\boldsymbol{A}^{cross}_{l,h}\in \mathbb{R}^{N_t \times N^{l}_{v}}\) in head \(h\) of layer \(l\), where \(N^{l}_{v}\) is the visual token number in layer \(l\), the confidence is calculated by:

\begin{equation}
    \boldsymbol{C}^h_l=\sum_{i=1}^{|\mathcal{T}|}\mathrm{Variance}(\boldsymbol{A}^{cross}_{l,h, t_i}),\, \mathrm{where}\, t_i \in \mathcal{T}.
\end{equation}
Heads with top \(N_{vis}\) confidence scores in layer \(l\) are considered visual heads, and indices are
\begin{equation}
    \mathcal{V}_{l} = \mathrm{TopK}(\boldsymbol{C}_l, N_{vis}).
\end{equation}

Once visual heads within the pruning layer are obtained, pruning is executed. At the pruning layer \(l\), we first aggregate cross-modal attention maps of the visual heads. The process is formulated as

\begin{equation}
    \boldsymbol{S}^{vis} = \frac{1}{|\mathcal{V}_{l}|} \sum_{j\in  \mathcal{V}_{l}}^{} \boldsymbol{A}^{cross}_{l,j},
\end{equation}
and the importance of a visual token is obtained by
\begin{equation}
    \boldsymbol{v}_{i} =  \frac{1}{N_t}  \sum_{j=1}^{N_t} \boldsymbol{S}^{vis}_{j,i}.
\end{equation}
With the target number \(N_2\) of retained visual tokens at the pruning layer, visual tokens corresponding to the top \(N_2\) scores in \(\boldsymbol{v} \in \mathbb{R}^{1 \times N_1}\) are preserved.

%% file: sections/5_evaluation.tex
\begin{table*}[t] 
\centering
\setlength{\aboverulesep}{0pt} %
\setlength{\belowrulesep}{0pt} %
\small
\begin{tabular}{l|ccccccc|c}
\toprule
\textbf{Method} & \textbf{GQA} & \textbf{MMB} & \textbf{MMB\(_{\mathrm{CN}}\)} & \textbf{MME} & \textbf{POPE} & \textbf{SQA}  & \textbf{VQA\(_{\mathrm{Text}}\)} &  \textbf{Average}  \\
\midrule
\rowcolor{lightgrays} LLaVA-1.5-7B & \multicolumn{8}{c}{\textit{Original 576 Tokens }} \\ 
Vanilla  & 61.9 & 64.7 & 58.1 & 1,862 & 85.9 & 69.5 & 58.2 & 100\% \\
\midrule
\rowcolor{lightgrays} LLaVA-1.5-7B &  \multicolumn{8}{c}{\textit{Retain 192 Tokens (\textbf{\(\downarrow\) 66.7\%})}} \\

FastV \tiny(ECCV24) & 52.7 & 61.2 & 57.0 & 1,612 & 64.8 & 67.3  & 52.5  & 89.5\%  \\

LLaVA-PruMerge \tiny(ICCV25) & 54.3 & 59.6 & 52.9 & 1,632 & 71.3 & 67.9  & 54.3  & 90.3\% \\

MustDrop \tiny(2024.11) & 58.2 & 62.3 & 55.8 & 1,787 & 82.6 & 69.2 & 56.5 & 96.5\% \\

PDrop \tiny(CVPR25) & 57.1 & 63.2 & 56.8  & 1,766 & 82.3 & 68.8 & 56.1 & 96.2\%  \\
HiRED \tiny(AAAI25) & 58.7 & 62.8 & 54.7 & 1,737 & 82.8 & 68.4  & 47.4  & 93.6\% \\
VisionZip \tiny(CVPR25) & 59.3 & 64.5 &57.3 &1,767 & 86.4 & 68.9 & 57.3 &98.2\% \\ 
SparseVLM \tiny(ICML25)  & 57.6 & 62.5 & 53.7 & 1,721 & 83.6 & 69.1  & 56.1 & 95.4\% \\
DART \tiny(EMNLP25)  & 58.9 & 63.6 & 57.0 & 1,856 & 82.8 & 69.8 & 57.4 &  98.1\% \\ 
HoloV \tiny(NIPS25) & 59.0 & 65.4 &	58.0&	1,820&	85.6&	69.8&	57.4 &98.9\%
 \\ 
ApET \tiny(CVPR26) & 60.2&	63.4&	57.9&	1,808	&86.3&	68.5&	54.4 & 97.8\% \\
 
\rowcolor[HTML]{F3EFFF}
\name{} \tiny(Ours)  & 60.9&	64.8&	59.1&	1,824	&85.5&	68.7&	57.4&	\textbf{99.3\%}
 \\
\midrule
\rowcolor{lightgrays} LLaVA-1.5-7B & \multicolumn{8}{c}{\textit{Retain 128 Tokens (\textbf{\(\downarrow\) 77.8\%})}} \\

FastV \tiny(ECCV24) & 49.6 & 56.1 & 56.4 & 1,490 & 59.6 & 60.2  & 50.6 & 83.8\%   \\

LLaVA-PruMerge \tiny(ICCV25) & 53.3 & 58.1 & 51.7  & 1,554 & 67.2 & 67.1 & 54.3 & 88.1\%   \\

MustDrop \tiny(2024.11) & 56.9 & 61.1 & 55.2 & 1,745 & 78.7 & 68.5 & 56.3 & 94.6\% \\

PDrop \tiny(CVPR25) & 56.0 & 61.1 & 56.6 & 1,644 & 82.3 & 68.3  & 55.1 & 94.2\% \\
HiRED \tiny(AAAI25) & 57.2 & 61.5 & 53.6  & 1,710 & 79.8 & 68.1 & 46.1 & 91.7\% \\
VisionZip \tiny(CVPR25) &57.6	&63.4	&56.7	&1,768&	84.7	&68.8	&56.8&	97.0\%
\\ 
SparseVLM \tiny(ICML25)  & 56.0 & 60.0 & 51.1 & 1,696 & 80.5 & 67.1  & 54.9 & 92.4\% \\
DART \tiny(EMNLP25)  & 57.9 & 63.2 & 57.0  & 1,845 & 80.1 & 69.1 & 56.4 & 96.8\% \\
HoloV \tiny(NIPS25)& 57.7	&63.9	&56.5&	1,802&	84.0&	69.8&	56.8	&97.4\%
\\  
ApET \tiny(CVPR26) &58.9	 &62.3&	56.4	&1,801	&86.1&	68.7	&53.9& 96.7\% \\
\rowcolor[HTML]{F3EFFF}
\name{} \tiny(Ours)& 60.2&	63.9&	57.8&	1,785&	86.3&	69.2	&56.7&	\textbf{98.4\%}
 \\
\midrule
\rowcolor{lightgrays} LLaVA-1.5-7B & \multicolumn{8}{c}{\textit{Retain 64 Tokens (\textbf{\(\downarrow\) 88.9\%})}} \\

FastV \tiny(ECCV24) & 46.1 & 48.0 & 52.7  & 1,256 & 48.0 & 51.1 & 47.8 & 74.1\%  \\

LLaVA-PruMerge \tiny(ICCV25) & 51.9 & 55.3 & 49.1  & 1,549 & 65.3 & 68.1 & 54.0 & 86.3\%   \\

MustDrop \tiny(2024.11) & 53.1 & 60.0 & 53.1 & 1,612 & 68 & 63.4 & 54.2 & 88.6\% \\ 

PDrop \tiny(CVPR25) & 41.9 & 33.3 & 50.5 & 1,092 & 55.9 & 68.6 & 45.9 & 72.5\% \\
HiRED \tiny(AAAI25) & 54.6 & 60.2 & 51.4 & 1,599 & 73.6 & 68.2 & 44.2 & 87.9\% \\
VisionZip \tiny(CVPR25) & 55.1&	60.1&	55.4&	1,690&	77.0&	69.0	&55.5&	93.2\%
 \\ 
SparseVLM \tiny(ICML25)  & 52.7 & 56.2 & 46.1  & 1,505 & 75.1 & 62.2 & 51.8 & 85.4\%  \\
DART \tiny(EMNLP25)  & 55.9 & 60.6 & 53.2 & 1,765 & 73.9 & 69.8 & 54.4 & 92.9\% \\
HoloV \tiny(NIPS25) & 55.3	&63.3	&55.1	&1,715	&80.3	&69.5	&55.4	&94.7\%
 \\ 

ApET \tiny(CVPR26) &56.9&	61.2&	54.4&	1,714	&84.4	&68.9&	53.0& 94.4\%
 \\ 
 \rowcolor[HTML]{F3EFFF}
\name{} \tiny(Ours)& 58.7&	62.3&	54.5&	1,723&	85.8&	69.4&	54.9&	\textbf{95.9\%}
 \\
\bottomrule
\end{tabular}
\caption{Task performance on LLaVA-1.5-7B.}
\label{llava-1.5}
\end{table*}

\section{Evaluation}

\subsection{Setup}

To evaluate \name{}, experiments are conducted across diverse benchmarks: GQA (\citealp{hudson2019gqa}), MMB (\citealp{liu2024mmbench}), MMB\(_{\mathrm{CN}}\) (\citealp{liu2024mmbench}), MME (\citealp{fu2023mme}), POPE (\citealp{li2023evaluating}), SQA (\citealp{lu2022learn}), and VQA\(_{\mathrm{Text}}\) (\citealp{singh2019towards}). We benchmark \name{} against a suite of visual token pruning baselines, including FastV (\citealp{chen2024image}), LLaVA-PruMerge (\citealp{shang2025llava}), MustDrop (\citealp{liu2024multi}), PDrop (\citealp{xing2024pyramiddrop}), HiRED (\citealp{arif2025hired}), VisionZip (\citealp{yang2025visionzip}), SparseVLM (\citealp{zhang2025sparsevlm}), DART (\citealp{wen2025stop}), HoloV (\citealp{zoudon}), and ApET (\citealp{ma2026apet}). Two models from the LLaVA family (\citealp{liu2023visual, li2024llava}), LLaVA-1.5-7B (\citealp{liu2024improved}) and LLaVA-NEXT-7B (\citealp{liu2024llavanext}), are used to verify the effectiveness of \name{}. To further validate the generalization, evaluation is performed on Qwen2.5-VL-7B (\citealp{Qwen2.5-VL}). The number of visual heads in each pruning layer of the above models is set to 6. For both LLaVA models, the shallow and middle pruning layers in \name{} are the 7th and 15th layers, respectively. Since the base LLM in Qwen2.5-VL-7B has fewer layers, we proportionally adjust pruning layers to the 6th and 13th layers to maintain consistency in relative pruning depth.

\begin{table*}[t]
\centering
\setlength{\aboverulesep}{0pt} %
\setlength{\belowrulesep}{0pt} %
\small

\begin{tabular}{l|ccccccc|c}
\toprule
\textbf{Method} & \textbf{GQA} & \textbf{MMB} & \textbf{MMB\(_{\mathrm{CN}}\)} & \textbf{MME} & \textbf{POPE} & \textbf{SQA}  & \textbf{VQA\(_{\mathrm{Text}}\)} &  \textbf{Average}  \\
\midrule
\rowcolor{lightgrays} LLaVA-NEXT-7B & \multicolumn{8}{c}{\textit{Original Tokens}} \\ 
Vanilla  & 64.2&	67.4&	60.6	&1,851&	86.5&	70.1	&64.9 & 100\% \\
\midrule
\rowcolor{lightgrays} LLaVA-NEXT-7B & \multicolumn{8}{c}{\textit{Retain 320 Tokens}} \\

FastV \tiny(ECCV24)  & 55.9 &	61.6 &	51.9	 &1,661	 &71.7	 &62.8	 &55.7 &	87.4\%  \\
LLaVA-PruMerge \tiny(ICCV25)& 53.6&	61.3	&55.3&	1,534	&60.8&	66.4&	50.6	&84.5\%   \\

MustDrop \tiny(2024.11)& 57.3&	62.8&	55.1&	1,641&	82.1&	68.0 &	59.9&	92.3\% \\

PDrop \tiny(CVPR25)& 56.4	&63.4	&56.2	&1,663	&77.6&	67.5	&54.4&	90.6\% \\
HiRED \tiny(AAAI25) &59.3	&64.2	&55.9&	1,690	&83.3&	66.7	&58.8&	93.3\% \\
SparseVLM \tiny(ICML25) & 56.1&	60.6	&54.5	&1,533	&82.4&	66.1&	58.4	&89.9\%  \\
DART \tiny(EMNLP25)&  61.7&	65.3	&58.2&	1,710&	84.1&	68.4	&58.7	&95.2\% \\
HoloV \tiny(NIPS25)& 61.7&	65.3	&57.5&	1,738&	83.9	&68.9	&58.7	&95.4\%
\\  

ApET \tiny(CVPR26) &61.0 &	63.5	&56.6&	1,783&	85.6&	69.7
	&54.4 & 94.4\% \\
\rowcolor[HTML]{F3EFFF}
\name{} \tiny(Ours)& 60.8&	65.9& 59.0 &	1,769&	86.1&	69.0&	57.1&		\textbf{95.9\%}
 \\
\bottomrule
\end{tabular}
\caption{Task performance on LLaVA-NEXT-7B.}
\label{LLAVA1.6}
\end{table*}

\subsection{Main Results} \label{Main}

We first evaluate \name{} across a range of benchmarks using the representative VLM LLaVA-1.5-7B. To provide a comprehensive assessment, we compare the task performance of both the baselines and our proposed framework under three pruning ratios: 66.7\%, 77.8\%, and 88.9\%. 

The experiment results are presented in Table \ref{llava-1.5}. Compared to the vanilla LLaVA-1.5-7B, applying \name{} leads to only a negligible performance drop of 0.7\% when 66.7\% of visual tokens are removed. Meanwhile, \name{} consistently outperforms all baselines in task performance across the three pruning ratios. Notably, its performance advantage becomes more pronounced as the pruning ratio increases. For instance, \name{} surpasses HoloV by 0.4\% at a pruning ratio of 66.7\%, and this margin increases to 1.2\% at 88.9\%, which highlights \name{}'s capability to preserve task-relevant visual tokens under more aggressive pruning.

\begin{table}[t]
\centering
\setlength{\aboverulesep}{0pt} %
\setlength{\belowrulesep}{0pt} %
\setlength{\tabcolsep}{2.6pt}
\small

\begin{tabular}{l|ccc c}
\toprule
\textbf{Method}  &  \makecell[c]{\textbf{Latency} \\ \textbf{(ms)}}
 &  \makecell[c]{\textbf{Prefilling} \\ \textbf{Time (ms)}}  & \makecell[c]{\textbf{Memory} \\ \textbf{Peak (GB)}} &  \makecell[c]{\textbf{FLOPs} \\ \textbf{(T)}}  \\
\midrule
\rowcolor{lightgrays} LLaVA-1.5-7B &  \multicolumn{4}{c}{\textit{Original 576 Tokens}} \\ 
Vanilla & 	204 &	127 & 19.01 & 4.24 \\
\midrule
\rowcolor{lightgrays} LLaVA-1.5-7B & \multicolumn{4}{c}{\textit{Retain 192 Tokens (\textbf{\(\downarrow\) 66.7\%}) }} \\

+SparseVLM  &	156 &	 69	 &  18.60	 &	1.76 \\
\rowcolor[HTML]{F3EFFF}
+\name{}   &  143 &	65 & 18.56	 &	1.66
 \\
 \midrule
\rowcolor{lightgrays} LLaVA-1.5-7B & \multicolumn{4}{c}{\textit{Retain 128 Tokens (\textbf{\(\downarrow\) 77.8\%})}} \\

+SparseVLM & 	147 &	69	 & 18.60 &	1.76 \\
\rowcolor[HTML]{F3EFFF}
+\name{}  &  133 	 &	54	 &  18.49	&	1.24
 \\
 \midrule
\rowcolor{lightgrays} LLaVA-1.5-7B & \multicolumn{4}{c}{\textit{Retain 64 Tokens (\textbf{\(\downarrow\) 88.9\%})}} \\
\midrule
+SparseVLM    &	145 &	69 & 18.60	 &	1.76  \\
\rowcolor[HTML]{F3EFFF}
+\name{} & 126	 &	48  & 18.43	&	0.82 \\
\bottomrule
\end{tabular}
\caption{Efficiency results on LLaVA-1.5-7B.}
\label{Efficiency}
\end{table}

\subsection{Results with High Resolution}

VLMs usually achieve better task performance under high-resolution visual inputs. However, this comes at the expense of a proliferation of visual tokens. To evaluate \name{} under high-resolution settings, we adopt LLaVA-NEXT-7B, which dynamically partitions each high-resolution image into multiple sub-images based on its original aspect ratio, producing up to 2,880 visual tokens per image. In our experiments, we set the number of retained visual tokens to 320, and the results are illustrated in Table \ref{LLAVA1.6}. Compared to the vanilla LLaVA-NEXT-7B, \name{} achieves an average task performance of 95.9\% with nearly negligible degradation on POPE. Moreover, among all pruning baselines, \name{} delivers the best overall performance, which outperforms ApET and DART by 1.5\% and 0.7\%, respectively.

\begin{table}[t]
\centering
\setlength{\aboverulesep}{0pt} %
\setlength{\belowrulesep}{0pt} %
\setlength{\tabcolsep}{2pt}
\small

\begin{tabular}{l|cccc|c}
\toprule
\textbf{Method} & \textbf{GQA}& \textbf{MMB} & \textbf{MME} & \textbf{POPE}    &  \textbf{Average}  \\
\midrule
\rowcolor{lightgrays} Qwen2.5-VL-7B &  \multicolumn{5}{c}{\textit{Original Tokens}} \\ 
Vanilla & 60.5&  83.3 &	2,327 &	86.2	& 100\% \\
\midrule
\rowcolor{lightgrays} Qwen2.5-VL-7B & \multicolumn{5}{c}{\textit{Token Pruning Ratio = 80\%}} \\

SparseVLM \tiny(ICML25)  & 54.7 & 76.0	&	2,063 &	73.6	  &	88.9\%  \\

PDrop  \tiny(CVPR25)  & 55.1  & 77.3 &	2,117 &	78.4		 &	91.5\%  \\

\rowcolor[HTML]{F3EFFF}
\name{} \tiny(Ours) & 55.1  & 76.0 &	2,092 &	87.0		 &	\textbf{93.3\%}
 \\
 \midrule
\rowcolor{lightgrays} Qwen2.5-VL-7B & \multicolumn{5}{c}{\textit{Token Pruning Ratio = 90\%}} \\

SparseVLM \tiny(ICML25)  & 51.3 & 71.7	 &	1,849 &	71.9	  &	83.5\%  \\

PDrop  \tiny(CVPR25)  & 52.0 & 73.6  &	1,886 &	74.8	&	85.6\%  \\

\rowcolor[HTML]{F3EFFF}
\name{} \tiny(Ours) & 52.9  &  72.7 &	1,950 &	83.4		&	\textbf{88.8\%}
 \\
\bottomrule
\end{tabular}
\caption{Task performance on Qwen2.5-VL-7B.}
\label{qwen}
\end{table}

\begin{table*}[t] 
\centering
\setlength{\aboverulesep}{0pt} %
\setlength{\belowrulesep}{0pt} %
\small

\begin{tabular}{l|ccccccc|c}
\toprule
\textbf{Method} & \textbf{GQA} & \textbf{MMB} & \textbf{MMB\(_{\mathrm{CN}}\)} & \textbf{MME} & \textbf{POPE} & \textbf{SQA}  & \textbf{VQA\(_{\mathrm{Text}}\)} &  \textbf{Average}  \\
\midrule
\rowcolor{lightgrays} LLaVA-1.5-7B & \multicolumn{8}{c}{\textit{Original 576 Tokens }} \\ 
Vanilla  & 61.9 & 64.7 & 58.1 & 1,862 & 85.9 & 69.5 & 58.2 & 100\% \\
\midrule
\rowcolor{lightgrays} LLaVA-1.5-7B &  \multicolumn{8}{c}{\textit{ Tokens Retained after Pruning: 219 (after Shallow), 60 (after Middle)}} \\

\textit{All Attention Heads} & 60.3 & 65.1 & 59.1 & 1,842 & 84.7 & 68.4  & 57.9  & 99.3\%  \\

\rowcolor[HTML]{F3EFFF} \textit{Visual Heads Only} & 60.6&	64.9&	59.5&	1,838	&85.0&	68.2&	57.6&	\textbf{99.3\%}
 \\
\midrule

\rowcolor{lightgrays} LLaVA-1.5-7B & \multicolumn{8}{c}{\textit{Tokens Retained after Pruning: 146 (after Shallow), 40 (after Middle)}} \\

\textit{All Attention Heads} & 59.3 & 64.9 & 59.5 & 1,851 & 83.5 & 68.7  & 57.0 & 98.8\%   \\

\rowcolor[HTML]{F3EFFF} \textit{Visual Heads Only} & 60.0&	65.0&	59.6&	1,819&	84.1&	68.5	&57.1&	\textbf{98.9\%}
 \\

\midrule
\rowcolor{lightgrays} LLaVA-1.5-7B & \multicolumn{8}{c}{\textit{Tokens Retained after Pruning: 73 (after Shallow), 20 (after Middle)}} \\

\textit{All Attention Heads} & 57.3 & 64.2 & 58.1  & 1,738 & 81.1 & 70.2 & 55.3 & 96.5\%  \\

\rowcolor[HTML]{F3EFFF} \textit{Visual Heads Only} & 58.2&	64.6&	58.2&	1,801&	82.9&	69.6&	54.8&	\textbf{97.4\%}
 \\

\midrule
\rowcolor{lightgrays} LLaVA-1.5-7B & \multicolumn{8}{c}{\textit{Tokens Retained after Pruning: 54 (after Shallow), 15 (after Middle)}} \\

\textit{All Attention Heads} & 56.2 & 63.6 & 57.7  & 1,724 & 79.9 & 69.1 & 54.6 & 95.3\%  \\

\rowcolor[HTML]{F3EFFF} \textit{Visual Heads Only} & 57.0&	64.3&	57.7&	1,745&	82.1&	69.8&	54.0&	\textbf{96.2\%}
 \\

\midrule
\rowcolor{lightgrays} LLaVA-1.5-7B & \multicolumn{8}{c}{\textit{Tokens Retained after Pruning: 36 (after Shallow), 10 (after Middle)}} \\

\textit{All Attention Heads} & 54.5 & 61.9& 55.1  & 1,658 & 77.2 & 68.4 & 53.0 & 92.4\%  \\

\rowcolor[HTML]{F3EFFF} \textit{Visual Heads Only} & 56.0&	62.8&	56.7&	1,709&	80.8&	68.9&	53.1&	\textbf{94.5\%}
 \\

\midrule
\rowcolor{lightgrays} LLaVA-1.5-7B & \multicolumn{8}{c}{\textit{Tokens Retained after Pruning: 18 (after Shallow), 5 (after Middle)}} \\

\textit{All Attention Heads} & 49.2 & 55.1 & 48.1  & 1,453 & 65.5 & 69.4 & 49.9 & 83.9\%  \\

\rowcolor[HTML]{F3EFFF} \textit{Visual Heads Only} & 52.4&	59.3&	51.5&	1,594&	75.7&	68.7&	50.5&	\textbf{89.2\%}
 \\

\bottomrule
\end{tabular}
\caption{Task performance on LLaVA-1.5-7B with pruning before the LLM backbone disabled under two shallow-layer pruning configurations: \textit{Visual Heads Only} and \textit{All Attention Heads}.}
\label{ablation}
\end{table*}

\subsection{Efficiency Analysis}

Efficiency of \name{} is evaluated on LLaVA-1.5-7B under the same pruning ratios in Section \ref{Main}. Using a single NVIDIA A40 (40 GB) GPU, we measure the inference latency, prefilling time, peak memory usage, and FLOPs on POPE. As shown in Table \ref{Efficiency}, when retaining 64 tokens, \name{} achieves an inference latency of 126 ms, representing a 38.2\% reduction compared to that of the vanilla LLaVA-1.5-7B. Moreover, the prefilling time is reduced by 62.2\%, indicating that \name{} effectively mitigates visual token redundancy, thereby reducing memory and computational overhead to accelerate inference. Compared to SparseVLM, \name{} consistently achieves superior efficiency across all evaluation metrics under various pruning ratios in Table \ref{Efficiency}. Importantly, as reported in Table \ref{llava-1.5}, \name{} also surpasses SparseVLM in performance. Taken together, these findings effectively indicate that \name{} not only enhances inference efficiency but also maintains competitive task performance.

\subsection{Results of Generalization}

To evaluate the generalization capability of \name{}, we further apply it to Qwen2.5-VL-7B, which features a different architecture from LLaVA models. The token pruning ratios are set to 20\% and 10\%, respectively. As illustrated in Table \ref{qwen}, \name{} consistently outperforms the baselines, which highlights its generalization. At a 10\% pruning ratio, it achieves a 3.2\% average improvement in task performance compared to PDrop. Notably, under a 20\% pruning ratio, \name{} attains 87.0\% on POPE, surpassing the vanilla Qwen2.5-VL-7B and indicating its capability to retain critical visual tokens.

\subsection{Ablation Study}

Ablation experiments are conducted on LLaVA-1.5-7B to assess the impact of shallow-layer pruning guided by \textit{Visual Heads Only} (additional ablation results to exhibit the effectiveness of \name{} are provided in Section~\ref{ablation_more}). Since pruning before the base LLM precedes shallow-layer pruning, it removes a portion of redundant tokens in advance, which may obscure potential deficiencies in the selection capability of shallow-layer pruning. To rigorously evaluate shallow-layer pruning, we disable the pruning before the base LLM and then compare performance between the original shallow-layer pruning and a modified variant guided by \textit{All Attention Heads}. Moreover, the pruning ratios are extended to more aggressive levels, demanding more precise token selection to preserve performance.

As presented in Table \ref{ablation}, the original shallow-layer pruning consistently achieves the best average performance across all pruning ratios. As the pruning ratio increases, the performance gap between this method and the \textit{All Attention Heads} variant gradually increases, growing from 0\% under mild pruning to as much as 5.3\% under more aggressive settings. This trend suggests that the advantage of \textit{Visual Heads Only} becomes increasingly pronounced when the model operates under stricter scenarios. Notably, under extreme pruning settings, it outperforms the variant on most, and in some cases all, evaluated benchmarks. These results collectively indicate that the original shallow-layer pruning effectively preserves critical visual tokens.

%% file: sections/6_conclusion.tex
\section{Conclusion}

Conventional visual token pruning methods estimate token importance by aggregating attention scores across all heads within the pruning layer in the LLM backbone. Our analysis reveals that such all-head aggregation may introduce redundant or even detrimental noise, ultimately degrading task performance. To address this issue, we propose \name{}, a training-free progressive visual token pruning framework that selectively utilizes a subset of heads for more precise token importance estimation. Experiments reveal that \name{} effectively improves inference efficiency while maintaining task performance, particularly under aggressive pruning ratios. These results highlight the importance of head-aware token importance estimation and provide new insights into efficient VLMs.